\documentclass[runningheads]{llncs}
\usepackage[T1]{fontenc}
\usepackage{multirow}
\usepackage{tikz}
\usepackage{hyperref} 
\usepackage{booktabs}

\usepackage{pifont}
\usepackage{amsmath}
\usepackage{cleveref}
\usepackage{svg}
\usepackage{orcidlink}
\usepackage{graphicx,verbatim}
\usepackage{subcaption}
\begin{document}
\title{Cross-Modal Contrastive Learning for the Retrieval of Immunotherapy-Associated Molecular Signatures from Histopathology}
\titlerunning{Cross-Modal Contrastive Learning for Immunotherapy}
%
\author{Sigrid Vila-Bagaria\inst{1}\orcidlink{0009-0002-9348-7235} \and
Mar Teixidó\inst{2}\orcidlink{0009-0002-0759-5644} \and Miquel Piñol\inst{2}\orcidlink{0000-0002-2073-0241
}  \and Felip Vilardell\inst{2}\orcidlink{0000-0002-0876-5425} \and
Robert Montal \inst{2}\orcidlink{0000-0002-8455-4232} \and
Veronica Vilaplana \inst{1}\orcidlink{0000-0001-6924-9961}}
\authorrunning{S. Vila-Bagaria et al.}
%
\institute{Signal Theory and Communications Department (TSC) \\
Universitat Politècnica de Catalunya - BarcelonaTech (UPC)\\
\email{\{sigrid.vila,veronica.vilaplana\}@upc.edu}\\ \and
Cancer Biomarkers Research Group (GReBiC) - IRB Lleida\\
\email{\{mteixido,rmontal\}@irblleida.cat}}

\maketitle              
\begin{abstract}
Gastric Adenocarcinoma is a leading cause of cancer mortality. Although ``Inflamed/Non-Inflamed'' subtypes have been proposed to predict immunotherapy response, their identification relies on a costly 10-gene RNA signature. We propose a Cross-modal Contrastive Multiple Instance Learning (CCMIL) framework for cross-modal retrieval, imputing these molecular signatures directly from standard Hematoxylin \& Eosin (H\&E) slides. By leveraging a supervised contrastive objective, CCMIL aligns visual morphological patterns with molecular phenotypes into a shared latent space. This establishes an interpretable search-by-case retrieval engine, enabling pathologists to query a whole slide image to surface transcriptomically coherent neighbors and approximate RNA signatures without genomic sequencing at inference. Our results demonstrate that this retrieval-first approach captures the continuous phenotypic spectrum of tumor inflammation and yields clinically interpretable attention heatmaps. 
Furthermore, the learned representation also supports competitive downstream classification, providing a practical molecular pre-screening strategy.


\end{abstract}

\section{Introduction} \label{sec:introduction}

Gastric adenocarcinoma (GAC) is a leading cause of cancer mortality globally \cite{sung2021global}, typically diagnosed at advanced stages with poor prognosis \cite{ajani2017gastric}. While immune checkpoint inhibitors (ICIs) have become a standard of care \cite{janjigian2023pembrolizumab,janjigian2021first,rha2023pembrolizumab}, heterogeneous responses demand better predictive biomarkers \cite{borcoman2019novel}. Recently, a distinct ``Inflamed/Non-Inflamed'' GAC phenotype, characterized by high immunogenicity, was identified as a strong predictor of ICI response \cite{rodriguez2025comprehensive}. However, determining this class requires a costly 10-gene RNA signature, limiting its routine clinical use. By learning to impute this molecular signature directly from low-cost, standard H\&E features, our framework bypasses the need for genomic sequencing at inference, providing a scalable pre-screening tool.

Deep learning (DL) in computational pathology offers an accessible alternative to molecular assays \cite{lu2021data}. However, learning the ``Inflamed/Non-Inflamed'' phenotype directly from H\&E faces two key challenges. First, paired histology-genomic datasets are scarce, increasing the risk of overfitting. Second, binarizing a continuous gene expression signature into a label introduces noise and discards the patient's true position along the biological phenotypic spectrum. To address these challenges, we investigate metric learning to build a highly interpretable search-by-case diagnostic tool. The contributions of this work are as follows.

\begin{itemize}
\item We propose a Cross-modal Contrastive Multiple Instance Learning (CCMIL) framework that learns a shared histology-transcriptomics embedding.
\item We demonstrate that the learned representation captures biologically coherent relationships between patients, enabling interpretable RNA signature imputation and competitive downstream classification from H\&E alone.
\end{itemize}


\section{Related Work} \label{sec:rel_work}

\textbf{Contrastive learning in computational pathology:} Contrastive learning optimizes latent spaces by pulling positive pairs together and pushing negative pairs apart \cite{hadsell2006dimensionality}. Frameworks like ConVIRT \cite{zhang2022contrastive} and CLIP \cite{radford2021learning} proved this highly effective for cross-modal retrieval in the vision-language domain. Medical adaptations, such as MedCLIP \cite{wang2022medclip}, apply these principles to healthcare, maximizing similarity between matched modalities (e.g., histology and genomics) to capture patient-specific biological associations. Crucially, establishing this tightly clustered joint manifold provides the foundational architecture required for robust retrieval tasks. 

\textbf{Cross-modal retrieval in computational pathology:} Accurate tumor subtyping is fundamental to precision oncology, yet unimodal MIL frameworks (e.g., ABMIL \cite{ilse2018attention}, CLAM \cite{lu2021data}, TransMIL \cite{shao2021transmil}) often fail to capture the full transcriptomic complexity of heterogeneous tumors. To address this, multimodal frameworks like MCAT \cite{chen2021multimodal} fuse histological and genomic data, but strictly require both modalities at inference, maintaining high sequencing costs. Emerging approaches instead leverage contrastive learning to align modalities \cite{li2025cross}, integrating spatial transcriptomics \cite{min2024multimodal} or enforcing semantic consistency \cite{yu2025contrastive} to enhance robustness. Building upon these advances, our framework exploits this aligned multimodal manifold specifically to enable cross-modal retrieval, allowing clinicians to impute molecular signatures using only histology at inference.


\usetikzlibrary{shapes.geometric, arrows.meta, positioning, calc, decorations.pathreplacing, backgrounds}


\definecolor{colorCold}{RGB}{77, 175, 74}     
\definecolor{colorColdRNA}{RGB}{166, 217, 164} 
\definecolor{colorHot}{RGB}{255, 127, 0}       
\definecolor{colorHotRNA}{RGB}{255, 197, 143}  

\begin{figure}[t]
    \centering
    \begin{subfigure}[b]{0.60\textwidth}
        \centering
        \includegraphics[width=\textwidth]{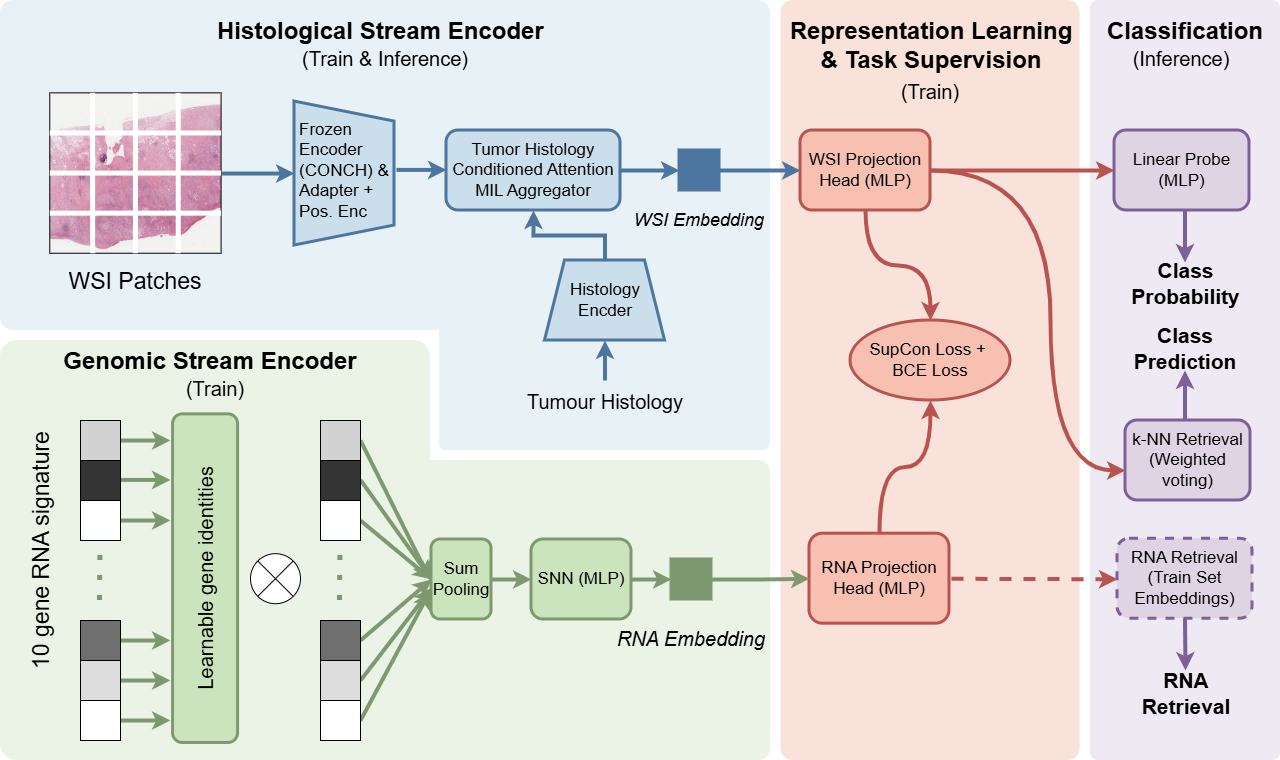}
        \caption{CCMIL Architecture}
        \label{fig:arquitectura}
    \end{subfigure}
    \hfill 
    \begin{subfigure}[b]{0.38\textwidth}
        \centering
    \resizebox{\textwidth}{!}{
\begin{tikzpicture}[
    scale=1.2, 
    font=\sffamily\Large,
    slide/.style={
        rectangle, 
        draw=white, 
        line width=0.5pt,
        inner sep=0pt, 
        minimum size=8mm, 
        rounded corners=1pt
    },
    rna/.style={
        regular polygon, 
        regular polygon sides=3, 
        draw=none, 
        inner sep=0pt, 
        minimum size=8mm, 
        rotate=0
    },
    alignline/.style={
        dashed, 
        very thick, 
        opacity=1.0, 
        line cap=round
    },
    bgGrid/.style={draw=gray!20, step=1.0, dashed, thin}
]
    \pgfmathsetseed{123}

    \draw[bgGrid] (0,0) grid (10,8);
    
    \node[anchor=north east, gray!60] at (10,-0.3) {Latent Dimension 1};
    \node[anchor=south, gray!60, rotate=90] at (-0.4, 4.0) {Latent Dimension 2};

    \foreach \x/\y [count=\i] in {
        2.5/2.0, 3.2/1.2, 3.8/2.5, 2.2/3.2, 
        4.0/3.5, 4.5/1.8, 3.0/4.0, 1.8/2.5
    } {
        \coordinate (S\i) at (\x,\y);
        
        \pgfmathsetmacro{\randAngle}{rand*360}
        \pgfmathsetmacro{\randDist}{0.7 + 0.2*rand}
        
        \coordinate (R\i) at ($(\x,\y) + (\randAngle:\randDist)$);
        
        \draw[alignline, colorCold!80!black] (S\i) -- (R\i);
        
        \node[rna, fill=colorColdRNA, draw=colorCold] at (R\i) {};
        \node[slide, fill=colorCold] at (S\i) {};
        
        \fill[white!30!colorCold] ($(\x,\y)+(-0.05,0.05)$) circle (0.5mm);
    }

    \node[colorCold, font=\bfseries\LARGE] at (3.0, 0.4) {Non-Inflamed};

    \foreach \x/\y [count=\i] in {
        7.0/5.5, 7.7/4.7, 8.3/6.0, 6.7/6.7, 
        8.5/7.0, 9.0/5.3, 7.5/7.5, 6.3/6.0
    } {
        \coordinate (HS\i) at (\x,\y);
        
        \pgfmathsetmacro{\randAngle}{rand*360}
        \pgfmathsetmacro{\randDist}{0.7 + 0.2*rand}
        
        \coordinate (HR\i) at ($(\x,\y) + (\randAngle:\randDist)$);
        
        \draw[alignline, colorHot!80!black] (HS\i) -- (HR\i);
        
        \node[rna, fill=colorHotRNA, draw=colorHot] at (HR\i) {};
        \node[slide, fill=colorHot] (HotSlide\i) at (HS\i) {};
        
        \fill[white!30!colorHot] ($(\x,\y)+(-0.05,0.05)$) circle (0.5mm);
    }

    \node[colorHot, font=\bfseries\LARGE] at (5.5, 7.6) {Inflamed};

    \draw[{Latex[length=4mm]}-{Latex[length=4mm]}, line width=1.2mm, draw=gray!80] (4.5, 3.8) -- (6.0, 5.0);
    \node[anchor=north west, align=center, fill=white, draw=black, rounded corners, inner sep=4pt] at (0.2, 6.7) 
    {\textbf{Class Separation} \\ \textbf{\& Modality Alignment}\\(SupCon Loss + BCE Loss)};

    \node[draw=black, fill=white, rounded corners, anchor=south east, align=left, inner sep=6pt] at (9.8, 0.2) {
        \textbf{Legend}\\
        \tikz\node[slide, fill=colorHot, minimum size=4mm] {}; H\&E Slide\\
        \tikz\node[rna, fill=colorHotRNA, draw=colorHot, minimum size=4mm] {}; RNA Signature\\
        \tikz\draw[alignline, colorHot!80!black] (0,0) -- (0.7,0); Contrastive Pair
    };

\end{tikzpicture}

}\caption{Joint Latent Space}  \label{fig:visual_abstract}
    \end{subfigure}
    
    \caption{The CCMIL framework. (a) Dual-stream architecture aligning WSI and RNA features during training to enable inference from H\&E alone. (b) The composite objective aligns modalities and separates ``Inflamed/Non-Inflamed'' cases in the joint latent space.}
    \label{fig:main_diagram}
\end{figure}

\newcommand{\cmark}{\ding{51}}%
\newcommand{\xmark}{\ding{55}}%

\newcommand{\std}[1]{{\scriptsize$\pm$#1}}
\begin{table}[t]
\centering
\caption{Quantitative performance comparison of the proposed framework against state-of-the-art baselines. All values are reported as mean $\pm$ standard deviation across 5-fold cross-validation.}
\label{tab:combined_results}
\setlength{\tabcolsep}{5pt} 
\resizebox{\textwidth}{!}{%
\begin{tabular}{@{}l l l l l l l@{}}
\toprule
\multicolumn{7}{c}{\textbf{Retrieval}} \\
\midrule
& & \multicolumn{3}{c}{\textbf{Retrieval (Image $\to$ RNA)}} & \multicolumn{2}{c}{\textbf{k-NN Classification}} \\
\cmidrule(lr){3-5} \cmidrule(lr){6-7}
\textbf{Model} & \textbf{Q $\rightarrow$ Tgt} & \textbf{R@1} & \textbf{R@5} & \textbf{MRR} & \textbf{Acc.} & \textbf{F1} \\ 
\midrule
Baseline & Te $\rightarrow$ Tr  & -- & -- & .71 \std{.106} & .57 \std{.129} & .53 \std{.109} \\
Baseline & Te $\rightarrow$ Te  & .67 \std{.115} & .89 \std{.086} & .78 \std{.067} & .67 \std{.107} & .65 \std{.074} \\
\textbf{CCMIL} & Te $\rightarrow$ Tr & -- & -- & \textbf{.75} \std{0.049} & \textbf{.68} \std{.080} & \textbf{.61} \std{.138} \\
\textbf{CCMIL} & Te $\rightarrow$ Te  & \textbf{.72} \std{.085} & \textbf{.92} \std{.051} & \textbf{.80} \std{.067} & \textbf{.71} \std{.155} & \textbf{.67} \std{.133} \\
\midrule
\midrule
\multicolumn{7}{c}{\textbf{Classification}} \\
\midrule
\textbf{Modality} & \textbf{Model} & \textbf{Acc.} & \textbf{AUC} & \textbf{F1} & \textbf{Prec.} & \textbf{Rec.} \\ 
\midrule
\multirow{3}{*}{Unimodal} 
 & CLAM & \textbf{.73} \std{.035} & \textbf{.79} \std{.034} & .72 \std{.051} & .68 \std{.031} & .76 \std{.065} \\
 & ABMIL & .72 \std{.021} & .79 \std{.078} & .70 \std{.022} & .68 \std{.048} & .72 \std{.073} \\
 & TransMIL & .52 \std{.117} & .72 \std{.082} & .62 \std{0.027} & .51 \std{0.104} & .85 \std{.191} \\ 
\midrule
Multimod. & MCAT & .66 \std{0.065} & .68 \std{.072} & .66 \std{.101} & .61 \std{.076} & .72 \std{.231} \\ 
\midrule
Cross-mod. & \textbf{CCMIL} & .72 \std{.157} & .75 \std{.135} & \textbf{.76} \std{.112} & \textbf{.68} \std{.117} & \textbf{.92} \std{.083} \\ 
\bottomrule
\end{tabular}%
}
\end{table}

\begin{table}[t]
\centering
\caption{Ablation study of the framework evaluating the Tumour Histology Conditioned Attention (THCA) and loss combinations for Classification and Retrieval.}
\label{tab:ablation}
\setlength{\tabcolsep}{4pt} 
\begin{tabular}{@{}c c c l l l l l@{}}
\toprule
\multicolumn{3}{c}{\textbf{Components}} & \multicolumn{3}{c}{\textbf{Classification}} & \multicolumn{2}{c}{\textbf{Retrieval}} \\
\cmidrule(lr){1-3} \cmidrule(lr){4-6} \cmidrule(l){7-8}
\textbf{THCA} & \textbf{SupConL} & \textbf{ClassL} & \textbf{Acc.} & \textbf{AUC} & \textbf{F1} & \textbf{R@1} & \textbf{R@5} \\ 
\midrule
\cmark & \cmark & \xmark & .54 \std{.092} & .45 \std{.156} & .66 \std{.069} & .69 \std{.068} & .88 \std{.068} \\
\cmark & \xmark & \cmark & .68 \std{.103} & .68 \std{.073} & .72 \std{.076} & .63 \std{.119} & .84 \std{.068} \\
\xmark &  \cmark &  \cmark & .72 \std{.106} & .74 \std{.084} & .74 \std{.080} & .60 \std{.069} & .84 \std{.107} \\
\midrule
\cmark & \cmark & \cmark & \textbf{.72} \std{.157} & \textbf{.75} \std{.135} & \textbf{.76} \std{.112} & \textbf{.72} \std{.085} & \textbf{.92} \std{.051} \\ 
\bottomrule
\end{tabular} \label{tab:ablation}
\end{table}

\begin{figure}[t]
    \centering
    \begin{subfigure}[b]{0.35\linewidth}
        \centering
        \includegraphics[width=\textwidth]{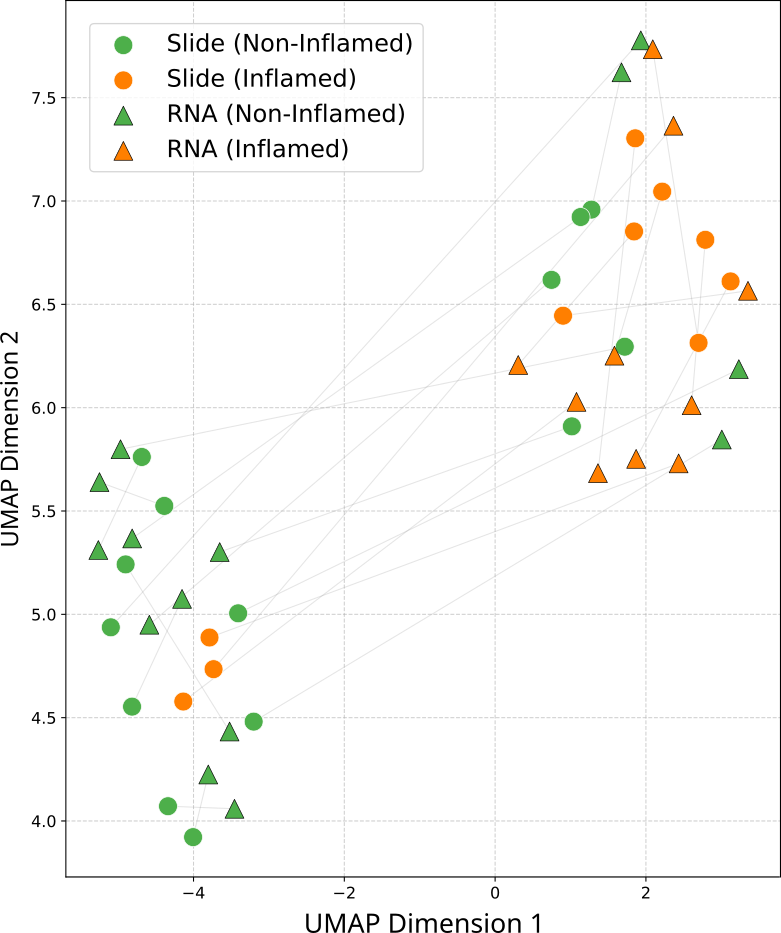}
        \caption{UMAP visualization of the joint latent space.}
        \label{fig:our_umap}
    \end{subfigure}
    \hfill
    \begin{subfigure}[b]{0.55\linewidth}
        \centering
        \includegraphics[width=\textwidth]{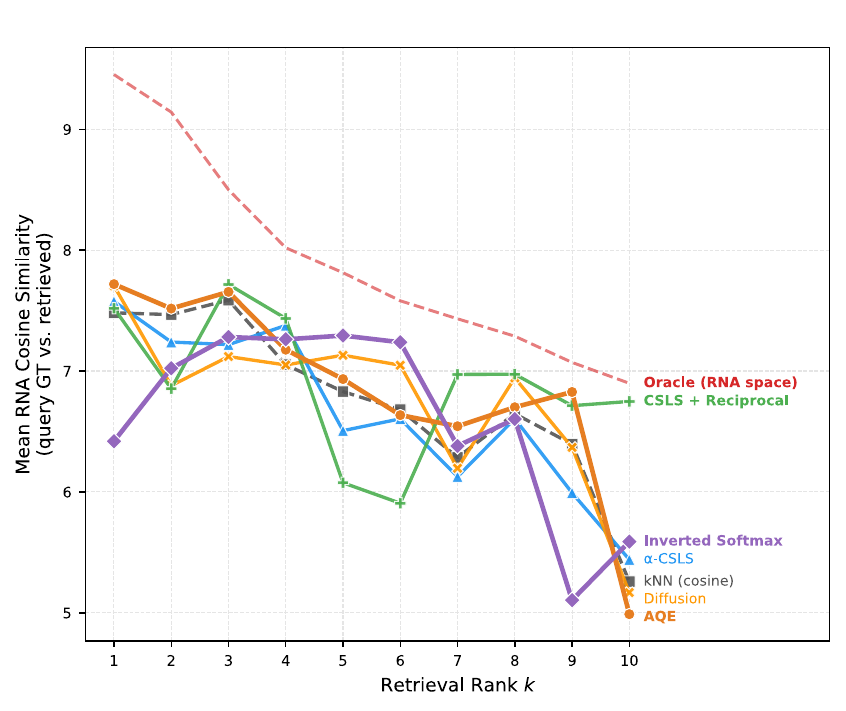}
        \caption{Mean RNA cosine similarity across top-$k$ retrieval ranks for cross-modal strategies.}
        \label{fig:retrieval_lineplot}
    \end{subfigure}

    \caption{Evaluation of the proposed framework. \textbf{(a)} Latent space visualization via UMAP demonstrating a structured joint representation. \textbf{(b)} Mean RNA Cosine Similarity between the query ground truth and retrieved cases across retrieval rank $k$. CSLS + Reciprocal strategy remains closest to the Oracle (retrieval in the RNA space) upper bound.}
    \label{fig:ccmil_evaluation}
\end{figure}

\begin{figure}[t]
    \centering
    
    \begin{subfigure}[b]{\linewidth}
        \centering
        \includegraphics[width=0.9\textwidth]{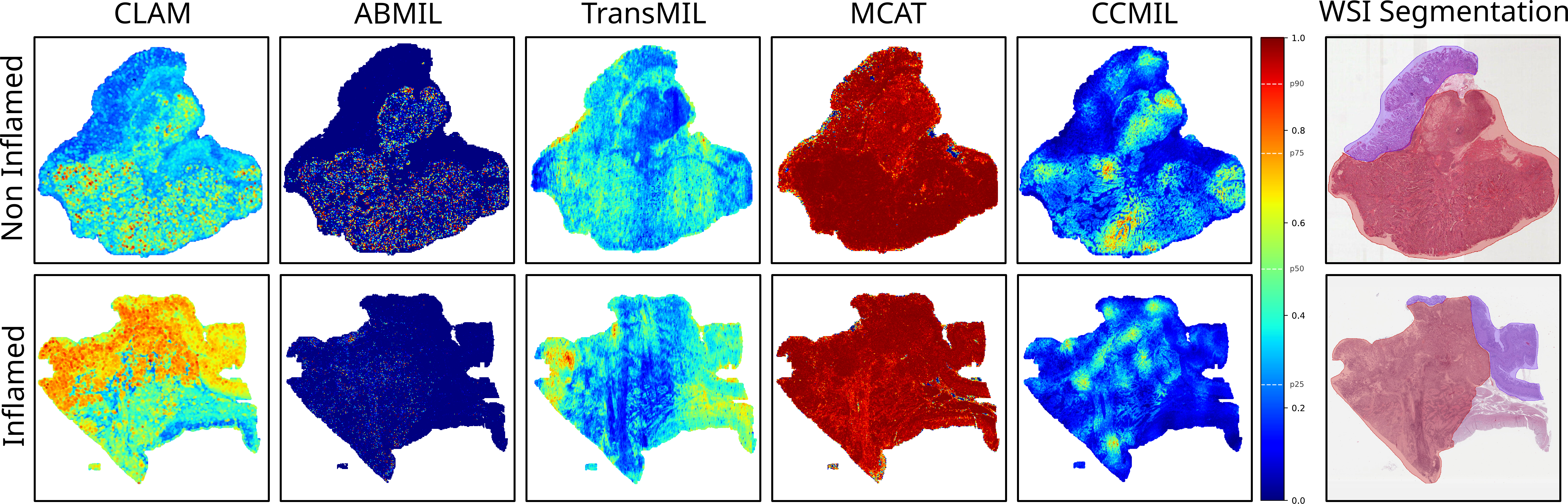}
        \caption{Spatial attention maps across models compared to ground-truth tissue segmentation (tumor in red, non-tumor in blue).}
        
        \label{fig:attention_heatmaps}
    \end{subfigure}
    
    \vspace{0.2cm} 
    
    \begin{subfigure}[b]{\linewidth}
        \centering
        \includegraphics[width=\textwidth]{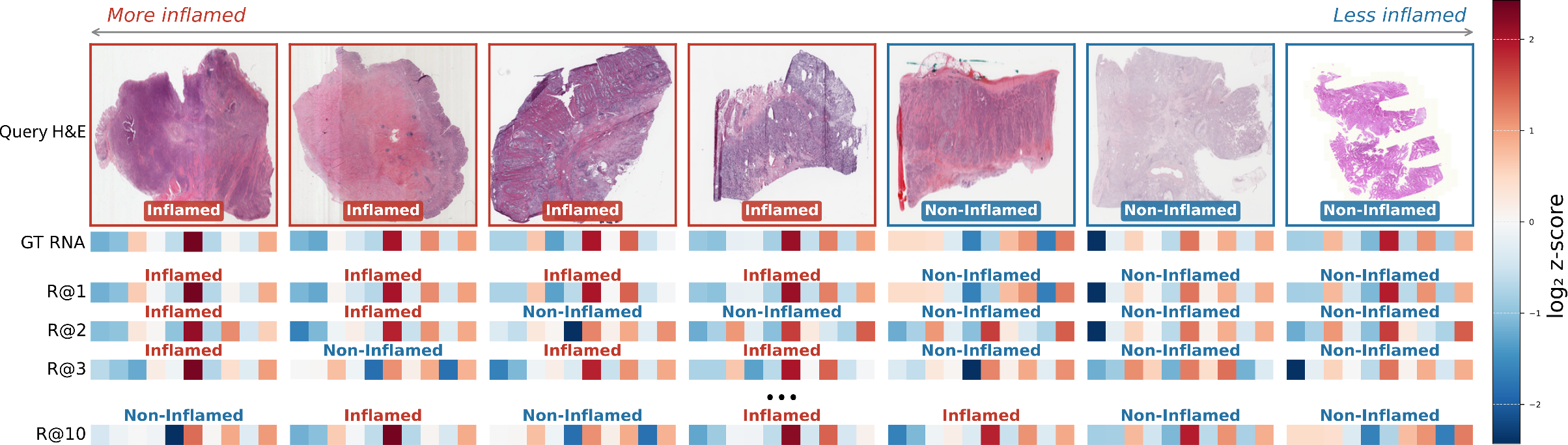}
        \caption{Continuous inflammatory spectrum: Test WSIs along the axis vector between class centroids, alongside their ground-truth and retrieved RNA expression profiles.}
        \label{fig:retrieval_progression}
    \end{subfigure}

    \caption{Qualitative evaluation of CCMIL. \textbf{(a)} CCMIL generates dense, highly localized attention hotspots compared to the baseline models. \textbf{(b)} The joint latent space captures a continuous inflammatory gradient, successfully retrieving transcriptomically coherent neighbors.}
    \label{fig:combined_qualitative}
\end{figure}

\section{Method} \label{sec:method}

\subsection{Cross-modal contrastive architecture}

The framework, as shown in Figure \ref{fig:arquitectura}, consists of two parallel encoding streams that map histological and genomic data to a shared joint embedding space.


\textbf{Histological stream:} Visual features extracted from the frozen CONCH backbone \cite{lu2024avisionlanguage} are refined by a visual adapter (a 2-layer MLP with layer normalization and ReLU) into a task-specific latent space. To preserve spatial structure within the permutation-invariant MIL setting, we employ Fourier Positional Encodings, mapping normalized patch coordinates to a high-dimensional frequency domain. Next, because different tumor histologies present distinct regions of interest, we compute the final slide-level embedding using our Tumor Histology Conditioned Attention (THCA), a lightweight gated-attention MIL module that conditions each patch's attention score on a learned histology-type embedding, yielding a histology-weighted sum of the adapted features. THCA guides the network to learn histology-specific attention profiles. Finally, an MLP projects these features into a shared space with the RNA embeddings, while a concurrently trained probe (2-layer MLP) enables direct WSI classification at inference.


\textbf{Genomic stream:} We encode the 10-gene expression signature into a latent representation utilizing a permutation-invariant embedding scheme followed by a Self-Normalizing Neural Network (SNN) \cite{klambauer2017self}. We assign a learnable identity vector to each gene. The initial aggregation is computed as the expression-weighted sum of these identities, since it learns a gene-specific semantic space while preserving permutation invariance. The aggregated vector is refined by an SNN composed of linear layers with Scaled Exponential Linear Units (SELU) and Alpha Dropout. Finally, similar to the histological stream, the features are projected to a common space with the RNA features using an MLP.

\subsection{Losses} \label{sec:loss}
To address label ambiguity and data scarcity, we optimize a composite objective to promote the latent space to be structured by molecular phenotype and separable for downstream classification.

\begin{equation}
    \mathcal{L}_{\text{total}} = \mathcal{L}_{\text{BCE}} + \mathcal{L}_{\text{SupCon}}
\end{equation}

\textbf{Representation Learning (Supervised Contrastive Loss):} We use a multi-modal Supervised Contrastive Loss ($\mathcal{L}_{\text{SupCon}}$) over the joint representations. Within a mini-batch of WSI and RNA embeddings, all features are first $L_2$ normalized and concatenated into a single set $I$ of size $2N$ (where $N$ is the batch size). For any anchor $i \in I$ with embedding $z_i$ and discrete label $y_i$, the set of positive keys $P(i)$ consists of all other samples in the batch sharing the same label ($p \in I \setminus \{i\} : y_p = y_i$). Because the WSI and RNA embeddings from the same patient inherently share the same label, this formulation naturally groups both \textit{instance-level cross-modal pairs} and \textit{class-level pairs} (any intra- or inter-modal embeddings with the same phenotype). The loss is formulated as:

\begin{equation}
    \mathcal{L}_{\text{SupCon}} = \frac{1}{2N} \sum_{i \in I} \frac{-1}{|P(i)|} \sum_{p \in P(i)} \log \frac{\exp(z_i \cdot z_p / \tau)}{\sum_{a \in I \setminus \{i\}} \exp(z_i \cdot z_a / \tau)}
\end{equation}

where $z_i \cdot z_p$ represents the cosine similarity between the embeddings, and the temperature parameter is $\tau = 0.07$. 

\textbf{Task Supervision (Classification Loss):} To enforce separability and refine the decision boundaries of the shared projection manifold, we apply a classification probe to both the aggregated WSI features and the genomic RNA features. The total classification loss is computed as the unweighted average of the Binary Cross-Entropy (BCE) losses from both modalities. This equal weighting encourages that both the histological and genomic representations independently maintain high diagnostic utility and are mapped correctly with respect to the target task's decision boundary.

\subsection{Inference strategy} 
At inference, the genomic stream is detached, enabling unimodal operation on standard H\&E WSIs to bypass sequencing costs while retaining multimodal benefits. The resulting embedding can be exploited in two complementary ways.

\textbf{Direct Classification:} The pre-trained probe projects the histological slide-level embedding onto the decision boundary to output a probability score.

\textbf{Retrieval Strategy:} Given a query H\&E slide, its embeddings are projected into the shared latent space and used to retrieve the most similar RNA embeddings from the reference dataset. We evaluated several cross-modal strategies: Inverted Softmax \cite{smith2017offline}, $\alpha$-CSLS \cite{lample2018word}, cosine $k$-NN, Diffusion \cite{iscen2017efficient}, and AQE \cite{chum2007total}. We found CSLS combined with Reciprocal reranking to provide the best retrieval performance, mitigating hubness and refining cross-domain alignment. By retrieving the top-$k$ ($k=5$) nearest neighbors using this strategy we can aggregate their corresponding RNA profiles to reliably impute a continuous molecular signature for the queried slide.
\section{Experimental setup} \label{sec:experimental}

\subsection{Dataset and preprocessing} \label{subsec:datasets}
We utilized a curated subset of the TCGA Gastric Adenocarcinoma cohort comprising $N=265$ diagnostic WSIs, manually reviewed by expert pathologists to ensure optimal tissue quality. The dataset consists of 142 ``Non-Inflamed'' and 123 ``Inflamed'' cases, with ground truth labels derived from a 10-gene RNA signature (\textit{ADORA2A, BLM, CCL20, CD3E, CD5, CD74, GLS, HLA-E, ITGAL, RPL23})  \cite{rodriguez2025comprehensive}. Additionally, cases were stratified by expert pathologists according to the histology Lauren classification into 173 Intestinal, 65 Diffuse, and 16 Mixed subtypes. WSI processing was standardized using the Trident \cite{zhang2025standardizing} framework for tissue segmentation and non-overlapping patching at 40$\times$. Finally, patch-level embeddings were extracted using the CONCH \cite{Lu_2023_CVPR} foundation model.

\subsection{Training setup}
\textbf{Framework and hardware} All experiments were implemented using the PyTorch framework, leveraging the pytorch-metric-learning library for loss formulations. Training was accelerated on a single NVIDIA A40 GPU (48GB VRAM). Due to the computational overhead of processing dense WSI feature bags, we utilized a batch size of $B=8$ patients with gradient accumulation of 4 steps, which was the maximum capacity allowed by memory constraints.

\textbf{Optimization} The network was trained for 50 epochs per fold using a 5-fold cross-validation scheme with early-stopping. Each fold was composed of a train, validation and test sets of 80\%, 10\% and 10\% of the cohort respectively. We employed the Adam optimizer with a fixed learning rate of $5 \times 10^{-5}$ and the loss defined in Section \ref{sec:loss}. 

\textbf{Baselines}
We compare CCMIL against representative state-of-the-art unimodal (H\&E-only) and multimodal (H\&E + RNA) approaches. We evaluate three widely used unimodal MIL models: ABMIL \cite{ilse2018attention}, CLAM-SB \cite{lu2021data}, and TransMIL \cite{shao2021transmil}. For multimodal fusion, we compare against MCAT \cite{chen2021multimodal}, which explicitly co-attends and fuses WSI and genomic features. For fair comparison, all models utilize identical CONCH embeddings and are evaluated across the same 5-fold splits. To isolate the contribution of our class-aware supervised contrastive framework, we evaluate a retrieval baseline that utilizes the same spatial architecture but relies solely on instance-level alignment (CLIP loss \cite{radford2021learning}). This setup mirrors standard multimodal contrastive methods widely used in the field.

\section{Results} \label{sec:results}
\subsection{Quantitative Evaluation}

We first select the retrieval strategy by evaluating multiple metrics on the validation set. As demonstrated in Figure~\ref{fig:retrieval_lineplot}, CSLS + Reciprocal reranking mitigates high-dimensional hubness and maintains the highest RNA cosine similarity to the Oracle upper bound for reliable signature imputation. Using this retrieval strategy, the 5-fold cross-validation results (Table~\ref{tab:combined_results}) demonstrate the effectiveness of CCMIL as a cross-modal retrieval engine. We report performance across two retrieval settings: Test $\to$ Train, mimicking real-world clinical inference by querying a new test slide against an established set of training cases, and Test $\to$ Test, which retrieves entirely within the test set to accurately compute R@1 and R@5 scores (percentage of queries that retrieve a match in the top $k$ results). Compared with the CLIP baseline, our class-aware formulation achieves a Test $\to$ Test R@5 of 0.92 and MRR of 0.80. Aggregating these retrieved WSI via weighted $k$-NN voting in the Test $\to$ Train setting yields a classification Accuracy of 0.72 and an F1-score of 0.76.

This learned latent space also enables effective downstream classification. CLAM achieves comparable Accuracy (0.73) but lacks clinical balance, while TransMIL and multimodal MCAT yield poor Precision (0.51, 0.61) due to domain gaps. Conversely, CCMIL achieves the highest F1-score (0.76) and Recall (0.92) indicating improved sensitivity for ICI pre-screening. Our ablation study (Table~\ref{tab:ablation}) confirms our architectural choices: representation learning alone yields poor decision boundaries (Accuracy 0.54), omitting the contrastive signal degrades retrieval (R@1 0.63), and replacing THCA with standard attention impairs both fine-grained cross-modal retrieval (R@1 0.60) and overall F1-score (0.74).

\subsection{Qualitative Evaluation}
Spatial attention heatmaps (Fig.~\ref{fig:attention_heatmaps}) highlight the interpretability of our framework. Compared to the baseline methods, which exhibit sparse under-activation (ABMIL), severe attention saturation (MCAT), or diffuse noise (CLAM), CCMIL generates dense, localized hotspots that align closely with ground-truth tumor segmentations. Because the ``Inflamed'' signature is visually indistinguishable to pathologists, these biologically coherent activations suggest CCMIL captures subtle morphological patterns associated with the ``Inflamed'' phenotype.

Furthermore, we show that the embedding space captures a continuous phenotype (Fig.~\ref{fig:retrieval_progression}). By projecting test slides onto an inflammation axis (the vector between class centroids), we selected seven evenly-spaced query slides. Displaying these alongside their retrieved RNA expression profiles (via Slide $\rightarrow$ RNA CSLS+Reciprocal reranking at ranks 1, 2, 3, and 10) confirms that CCMIL organizes slides along a biologically meaningful gradient, surfacing transcriptomically coherent neighbors even at larger retrieval ranks.
\section{Discussion and Conclusions} \label{sec:discussion_and_conclusions}

We introduce CCMIL, a Cross-modal Contrastive MIL framework that enables interpretable retrieval of molecular signatures directly from standard H\&E slides. By integrating a composite loss with our THCA module, CCMIL learns a shared histology-transcriptomics latent space during training, improving cross-modal alignment relative to the evaluated baselines. Beyond binary classification, CCMIL acts as an interpretable search-by-case engine, allowing pathologists to retrieve transcriptomically coherent neighbors and impute a patient's continuous inflammatory phenotype.

This learned retrieval-oriented representation also supports effective downstream classification, yielding a highly sensitive classifier for immunotherapy pre-screening. While our cross-validation demonstrates consistent performance within the TCGA cohort, this proof-of-concept has been evaluated only on a retrospective cohort. Future work will focus on validating the framework in an independent external cohort to assess generalizability. Ultimately, by providing biologically meaningful spatial heatmaps and bypassing genomic sequencing at inference, CCMIL offers a scalable framework for precision immuno-oncology.

\begin{credits}
\subsubsection{\ackname} This work was supported by the research project PID2023-148614OB-I00, funded by MICIU/AEI/10.13039/501100011033 and by FEDER, EU. The research conducted by Sigrid Vila-Bagaria is supported by the AGAUR-FI predoctoral programme (2025 FI-STEP 00032) of the Secretariat for Universities and Research of the Department of Research and Universities of the Generalitat of Catalonia, and by the European Social Fund Plus. This study has been funded by Instituto de Salud Carlos III (ISCIII) through the research project “PI24/01017”, and co-funded by the European Union. MTM acknowledges the support from Diputació de Lleida – La força dels municipis and from the Intramural Research Program of Institut de Recerca Biomèdica de Lleida (IRBLleida). RM acknowledges the support from Instituto de Salud Carlos III (Juan Rodés contract: JR21/00039) co-funded by the European Union.

\subsubsection{\discintname} 
The authors have no competing interests to declare that are relevant to the content of this article.
\end{credits}

%
%
%
\bibliographystyle{splncs04}
\bibliography{bibliography}

\end{document}